\pdfoutput=1
\documentclass[sigconf,nonacm]{acmart}

\usepackage{booktabs}
\usepackage{multirow}
\usepackage{xspace}
\usepackage{enumitem}
\usepackage{tikz}
\usepackage{pgfplots}
\pgfplotsset{compat=1.18}
\usepackage{comment}
\usepackage{fontawesome5}
\usepackage[skins]{tcolorbox}
\definecolor{vouchblue}{HTML}{2F5D9A}
\newtcolorbox{rulebox}{enhanced, colback=white, frame hidden, sharp corners,
  borderline={0.4pt}{0pt}{black!35}, borderline north={1.6pt}{0pt}{vouchblue},
  left=4pt, right=4pt, top=1pt, bottom=1pt, before skip=2pt, after skip=2pt, fontupper=\footnotesize}
\usetikzlibrary{positioning,arrows.meta,fit,calc,shapes.misc}

\newcommand{\otf}{OTF-Tools\xspace}
\newcommand{\devs}{DEVS-Gen\xspace}

\newcommand{\gate}{Vouch\xspace}

\newcommand{\dterm}[1]{\textsf{#1}}
\newcommand{\Trusted}{\dterm{Trusted}\xspace}
\newcommand{\Caught}{\dterm{Caught}\xspace}
\newcommand{\Already}{\dterm{Already caught}\xspace}
\newcommand{\Untestable}{\dterm{Untestable}\xspace}
\newcommand{\Failhist}{\dterm{Failed history check}\xspace}
\newcommand{\Cantexpress}{\dterm{Can't express}\xspace}
\newcommand{\Badrequest}{\dterm{Request rejected}\xspace}

\newcommand{\sig}[1]{{\boldmath\bfseries #1}}

\newcommand{\rProgsClaude}{62}

\newcommand{\rGateCovClaude}{80.1}%
\newcommand{\rGateErrClaude}{0.1}%
\newcommand{\rGateWrongClaude}{5}

\newcommand{\rNoCheckCovClaude}{90.0}%
\newcommand{\rNoCheckErrClaude}{4.4}%
\newcommand{\rNoCheckWrongClaude}{277}

\newcommand{\rHistWrongClaude}{207}
\newcommand{\rWrongRemovedHistClaude}{97.6}%
\newcommand{\rRightLostClaude}{7.1}%
\newcommand{\rRightLostCountClaude}{430}
\newcommand{\rRandomErrClaude}{4.4}%
\newcommand{\rOtfHistWrong}{202}

\newcommand{\rItAlreadyRight}{71}%

\newcommand{\rProgsWithWrong}{19}%
\newcommand{\rProgsWrongKept}{4}%
\newcommand{\rOtfLargestProgWrong}{32}%

\newcommand{\rGateCovOllama}{4.2}

\newcommand{\rNoCheckErrOllama}{69.6}%

\newcommand{\rCaughtWrong}{44}
\newcommand{\rAlreadyWrong}{158}

\newcommand{\rUntestableRight}{43}%
\newcommand{\rUntestableRightNoEffect}{37}%
\newcommand{\rRouteProbWrong}{195}%

\newcommand{\rInterpWf}{240 of 240}

\newcommand{\rInterpMl}{209 of 216}
\newcommand{\rMlCovInterp}{79.9}%
\newcommand{\rMlCovOracle}{84.3}%

\newcommand{\rPolPNoCheckCov}{93.9}
\newcommand{\rPolPNoCheckErr}{47.1}%

\newcommand{\rPolPDisCov}{47.4}%

\newcommand{\rPolHGateCov}{72.1}%

\newcommand{\rPolHHistCov}{77.5}

\newcommand{\rAnsJudgeN}{2{,}028}
\newcommand{\rAnsJudgeErrClaude}{1.1}%
\newcommand{\rGateErrAnsSub}{0.2}%

\newcommand{\rReactErrClaude}{1.4}%
\newcommand{\rGateErrReactSub}{0.0}%

\newcommand{\rProgJudgeErrClaude}{3.5}%

\newcommand{\rSonnetAnsOwnCov}{94.1}%

\newcommand{\rJevN}{2{,}028}
\newcommand{\rJevErrClaude}{2.3}%

\newcommand{\rJevAUROC}{0.63}%
\newcommand{\rHaikuAnsAUROC}{0.86}
\newcommand{\rJevErrOtf}{6.0}%
\newcommand{\rJevLongRows}{410}%
\newcommand{\rJevDKeptOllama}{$-12.9$~[$-23.3$,~$-4.5$]}%
\newcommand{\rJevUSDPerRow}{0.0011}%
\newcommand{\rJevSecPerRow}{0.33}%

\newcommand{\rReactRanProgHaiku}{41.8}%
\newcommand{\rFreshRightRecovered}{71}%
\newcommand{\rFreshExecPerQ}{1.92}%

\newcommand{\rGateExecPerQ}{1.5}%
\newcommand{\rGateExecMedian}{1}
\newcommand{\rGateSecPerQ}{4.2}%
\newcommand{\rAnsJudgeUSDPerRow}{0.0037}%
\newcommand{\rAnsJudgeSecPerRow}{28.6}%

\newcommand{\rReactUSDPerRow}{0.0038}%
\newcommand{\rReactSecPerRow}{24.7}%

\newcommand{\rTotalUSD}{79}%

\begin{document}

\title{Reproducing History Is Not Enough: Intervention-Level Assurance for LLM-Generated Simulators}

\author{Ashraf Abdul-Muumin}
\affiliation{%
  \institution{University of Alabama}
  \city{Tuscaloosa}
  \state{Alabama}
  \country{USA}}
\email{aashraf@crimson.ua.edu}

\author{Aniruddha Chattopadhyay}
\affiliation{%
  \institution{PVX Partners}
  \country{India}}
\email{studyaniruddha@gmail.com}

\author{Noorbakhsh Amiri Golilarz}
\affiliation{%
  \institution{University of Alabama}
  \city{Tuscaloosa}
  \state{Alabama}
  \country{USA}}
\email{namirigolilarz@ua.edu}

\author{Kaushik Roy}
\authornote{Corresponding author.}
\affiliation{%
  \institution{University of Alabama}
  \city{Tuscaloosa}
  \state{Alabama}
  \country{USA}}
\email{kroy2@ua.edu}

\renewcommand{\shortauthors}{Abdul-Muumin et al.}

\begin{abstract}
Large language models can now write simulation programs of operational processes and workflows. These programs can be reused to answer what-if questions on new records and policies. However, a program that reproduces recorded history can still give wrong answers to what-if questions; requested changes may be applied wrongly, or they may be applied to records other than the ones the question is about, or the program may stop being valid once its policy is revised. Current checks return a pass-or-fail verdict for the whole program, which says little about whether any single answer can be trusted. We propose \emph{Vouch}, a library that judges whether an answer from a generated simulator can be trusted for the change being asked about. Each what-if question is turned into a structured change, which is checked against the program and the current records. The library also keeps a record of the kinds of change each program has passed or failed, and repeats the relevant tests whenever the program is reused. We apply it to three program-generation platforms, with programs built by five language models, across four domains: manufacturing, IT operations, and pharmacy and loan-application workflows, using independently checked reference answers. Our baselines are the platforms' own checks where such checks exist, an LLM-as-a-judge, and a ReAct agent that can run the program. We measure how many correct answers are trusted, how many are wrongly rejected, how many wrong answers are accepted, and what the checking costs.

On programs written by Haiku and Sonnet in one platform, Vouch removes \rWrongRemovedHistClaude\% of the wrong answers, while still trusting \rGateCovClaude\% of all answers. At equal coverage, it accepts fewer wrong answers than LLM judges built on Claude Haiku. We disclose code and results for reproducibility \href{https://anonymous.4open.science/r/vouch_artifact-6893}{\textcolor{ACMDarkBlue}{\faGlobe\,here}}.

\end{abstract}



\maketitle

\section{Introduction}
\label{sec:intro}

Large language models (LLMs) are now capable of writing programs that recompute processes with changes applied to answer what-if questions. Given a document that states a process's rules and its history records, a capable model can produce a simulator: code that reads the recorded inputs, applies the rules, and reports what happens~\cite{otftools2026,chen2026devs,holt2025gsim}. Once built, the same program can be used to answer questions about new records or revised policies. Thus, cost-wise, building such programs once is cheaper than having a model reason through the process for each question~\cite{otftools2026}.

This paper investigates the trustworthiness of these reusable programs. Platforms check the generated program before it is used. The usual check asks whether the program can reproduce the recorded history. The platform, in turn, hides some of the recorded outcomes, runs the program on the recorded inputs, and has the results compared with the hidden outcomes. This is also a standard test for a simulation model~\cite{sargent2013vv}. We call this a \emph{history check}. The check gives a pass-or-fail verdict for the whole program. If the program passes, it is accepted and can be reused for subsequent questions. A what-if question, however, asks about one particular change, and a program can reproduce history perfectly and \emph{still} handle that change wrongly.

\begin{figure*}[t]
  \centering
  \includegraphics[width=\textwidth]{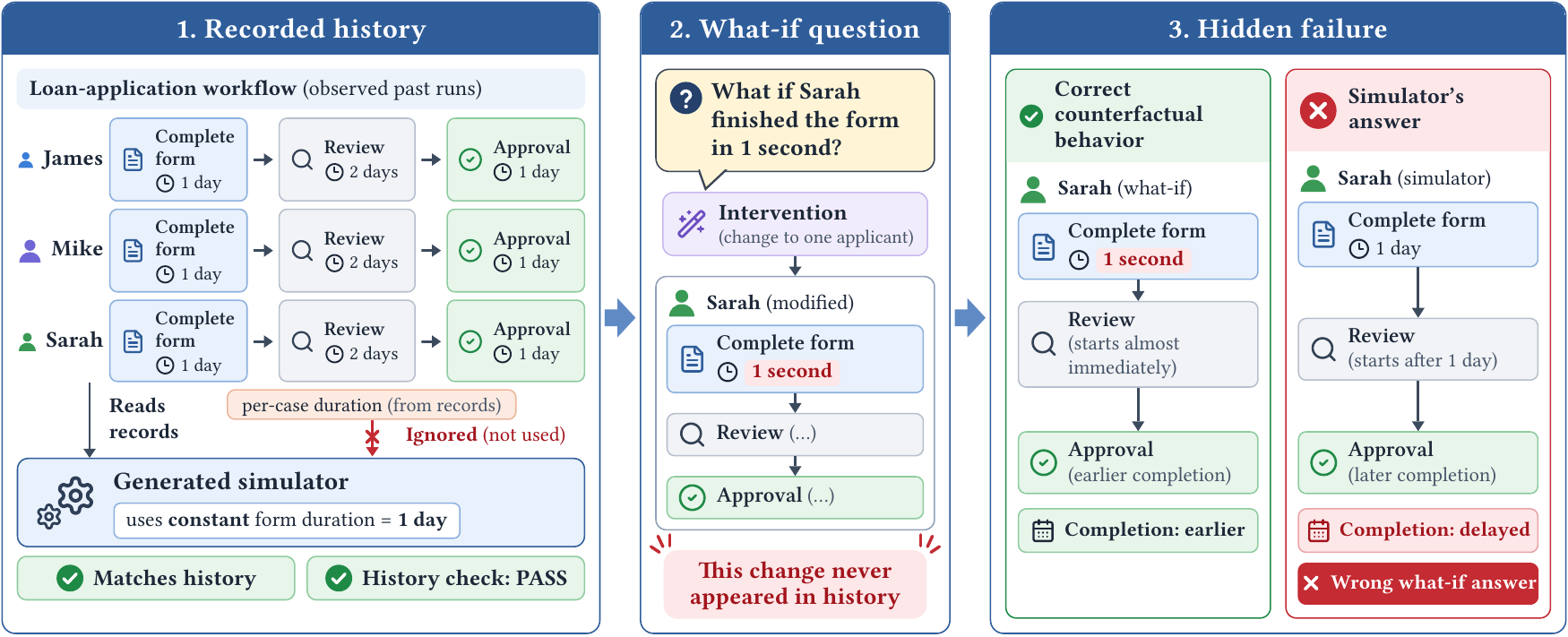}
  \caption{The gap in LLM-generated simulators built from historical records. (1) Every applicant in the records took one day on the form; the simulator ignores the per-case durations and assumes one day for everyone, yet still passes the history check. (2) Asked what if Sarah had finished the form in one second, it still applies one day and predicts a delayed finish. (3) Reproducing history cannot show that a program handles new interventions; this is the gap Vouch tests for.}
  \label{fig:history-gap}
\end{figure*}

Take, for example, a loan-application process from our benchmark in which every applicant spends exactly one day completing a form. A program written by one of the models reproduced every recorded outcome on all four runs, and the platform's own check accepted it with a perfect score. The program's code, however, showed a serious gap: it ignored the per-case durations it read from the records and used the published one-day constant instead, as shown in \textbf{Figure~\ref{fig:history-gap}}. That duration never varied in history, so no history check \emph{saw} the problem. The program had assumed every applicant took one day, so when asked what would happen if an applicant finished in one second, it answered a full day late. This gap also extended to other cases waiting for the same workers.

Machine learning has a name for this situation: models that fit the same data equally well can still behave differently on inputs the data never exercised~\cite{damour2022underspecification}. A single pass-or-fail verdict on a program therefore says little about whether a \emph{particular answer}, for a \emph{particular change}, can be trusted. Failures can stem from various places: \textbf{(a)} the question may be misread, \textbf{(b)} the change may be applied to the wrong record, \textbf{(c)} the program may ignore the changed input or simulate its effects incorrectly, or \textbf{(d)} the program may stop being valid when the policy it was built for changes. A check that gives one verdict for the whole program cannot tell these failures apart, and cannot tell the user which answers to rely on.

\paragraph{Our approach.} We built Vouch, an assurance library that decides, for each answer, whether to trust it or decline it, based on the change the question requests. Because the change is an intervention on the process, we call this intervention-level assurance. Vouch does this for each question:
\begin{enumerate}[leftmargin=*,nosep]
  \item turns the question into a structured change (what is changed, how, by how much, when, and for which record) and checks that change against the document and the current records;
  \item checks that the program reproduces the recorded history of the run being asked about;
  \item runs small tests aimed at the requested \emph{kind} of change, built only from the document's stated rules and records, never from reference answers; for example, it changes the same kind of input for another case and checks that the program's output moves, and that nothing before the change moves;
  \item keeps a record of which kinds of change each program has passed or failed, so that a failure blocks later answers that depend on the same kind of change, and repeats the relevant tests when the program is reused on new records or under a new policy.
\end{enumerate}
Every decline comes with a reason: either a test showed the program getting this kind of change wrong (`\textbf{\Caught}'), or it had failed the same test on an earlier question (`\textbf{\Already}'), or \emph{no} test of this kind of change could run on the program (`\textbf{\Untestable}'), or the program does not reproduce the recorded history (`\textbf{\Failhist}'), or lastly, it has no input that can represent the change (`\textbf{\Cantexpress}').

\paragraph{Evaluation.} We evaluate Vouch as one of several \emph{trust judges}, procedures that decide whether to trust an answer, all deciding about the same fixed programs, questions and answers. For the other judges, they are: \textbf{(a)} each platform's own acceptance check, \textbf{(b)} the history check by itself, \textbf{(c)} generic tests that are not aimed at the requested change, \textbf{(d)} agreement with an answer an LLM gives by reading the document, \textbf{(e)} an LLM asked to judge the answer, \textbf{(f)} an LLM asked to judge the program, and \textbf{(g)} a ReAct agent~\cite{yao2023react} that can run the program in a sandbox before deciding. Programs come from three platforms: \otf~\cite{otftools2026}, \devs~\cite{chen2026devs}, and an LLM-based coding agent, built by five LLMs, in four domains: manufacturing, IT operations, and loan and pharmacy workflows. Reference answers come from separately implemented and independently checked reference simulators, never from an LLM. Because a judge can always look safer by answering less, we compare judges at the \emph{same answer coverage}, and we report the cost of each.

\paragraph{Findings.} With no added checks, \rNoCheckWrongClaude{} of the answers given by programs generated by Claude's Haiku 5.5 and Sonnet 5.5 models are wrong (\rNoCheckErrClaude\%). The history check leaves \rHistWrongClaude{} of them. Vouch, our library, leaves \rGateWrongClaude{}, while trusting \rGateCovClaude\% of all program-question pairs. Almost all of the difference stems from one platform, where \rOtfHistWrong{} wrong answers pass the history check and Vouch trusts none of them, at the cost of 123 of the 2{,}052 correct answers the history check keeps. The faults were with the generated programs, as every wrong answer came from a correctly interpreted question. For scenarios where programs make no errors, or where the history check already find the errors, Vouch adds nothing. At equal coverage it accepts fewer wrong answers than LLM judges built on Claude Haiku, and ties with stronger and open-weight judges that we ran on small samples. When the rules change, here by adding one worker to a pool, every program fails the history check and Vouch answers nothing, while agreement with a reading answer keeps nearly half of the answers without error. Vouch needs no extra model calls and about \rGateExecPerQ{} extra program runs per question.

\paragraph{Contributions.}
\begin{itemize}[leftmargin=*,nosep]
  \item Vouch, a method for deciding whether to release a generated simulator's answer to a what-if question, which tests the requested kind of change without reference answers and names the reason for every decline (\S\ref{sec:method}).
  \item A benchmark and protocol for answer-level assurance: 547 test questions in four domains with exact answers under fixed random draws, independently checked reference answers, no-effect controls and policy variants, and 86 programs from three platforms and five builder models, judged with the candidate answers held fixed (\S\ref{sec:problem}, \S\ref{sec:benchmark}, \S\ref{sec:judges}).
  \item A characterization of when answer-level checking helps or falls short, relative to rule-based and LLM judges at equal coverage, reporting results across platforms, domains, change types, and generated programs (\S\ref{sec:results}).
\end{itemize}

\section{Problem Setting}
\label{sec:problem}

We now formalize the core components of our setting: the documents and runs describing a process, the programs derived from them, the questions posed to these programs, and lastly, the judges that determine which answers to release.

\paragraph{Documents, runs and cases.} A \textbf{document} describes a process, such as how a loan application moves from intake to approval. It begins by specifying the process rules, for example, the steps, worker or machine assignments, working hours, routing logic, and capacity sharing. This is then followed by one or more \textbf{runs}. A run represents a recorded execution period of the process, detailing both it's inputs and it's outcomes. Examples of inputs include, but are not limited to, arrivals, planned work, orders, and task requests. For outcomes, they may be events like case completion times, shipped items, or as simple as task start times. A \textbf{case} is one item that moves through the process: one loan application, one order, or one job on a cluster. A case may pass through the same activity more than once; each pass is a \textbf{visit}.

\paragraph{Programs.} Once a \emph{platform} reads a document, it produces a \textbf{program}. A program is the simulator that takes a run's inputs and returns its outcomes. The program is written by an LLM, the platform's \emph{builder model}. Most platforms check it before accepting it. Every program we build comes from development documents and is then applied, unchanged, to test runs it has never seen. This is the reusable attribute described in the introduction.

\paragraph{Questions.} A what-if question specifies a run, a single modification, and a target outcome to report. For example: ``In Run~11, suppose one Technician (Technician\_7) was unavailable from 8:00 to 12:00 on the first day. At what time would case c0014 end?'' Each modification belongs to a \emph{family}, a category of change such as altered activity durations, updated routing probabilities, or simply reduced workers (Table~\ref{tab:benchmark}). Notably, valid modifications do not always alter the queried outcome; if Technician\_7 never handled case c0014 or any cases preceeding it, the completion time remains unchanged. We deliberately include these \emph{no-effect controls}, because a judge that indiscriminately rejects any answer matching the recorded baseline would falsely decline them.

\paragraph{Answers.} To answer, the platform applies the change to the run's inputs, runs the program, and reads off the queried value. An answer is correct if it equals the reference answer, computed by an independently checked reference simulator (\S\ref{sec:benchmark}), to within one unit (one second or one item). A program may also give no answer.

\paragraph{Trust judges.} A \textbf{trust judge} inspects the program, the run, the question and, for answer-level judges, the candidate answer itself. The judge decides whether to accept the answer or abstain. The simplest baseline is the \emph{history check}, which trusts any answer by a program that reproduces the run's recorded outcomes exactly. Finally, an answer counts as \emph{released} only if the program produces a valid output and the judge accepts it.

\paragraph{What we measure.} We evaluate judges across program--question pairs. We report four primary metrics: \textbf{(1)} coverage, the fraction of all pairs whose answer is released; \textbf{(2)} error, the share and count of released answers that are wrong; \textbf{(3)} correct answers kept, the fraction of all pairs that are both released and correct; \textbf{(4)} cost, measured in program executions, model calls, dollars, and time. Due to the fact that a judge that abstains on every input achieves essentially \emph{zero errors} without utility, we evaluate methods at equal coverage. Confidence-scoring judges are thresholded at Vouch's coverage to retain their most confident outputs, whereas binary judges are assessed at their native coverage alongside a \textbf{random-refusal} baseline matched to Vouch's coverage. We match coverage separately per domain, platform, and builder model, never pooling Claude-generated programs with open-weight programs, whose error rates are ten times higher.

\section{Our Method: Vouch}
\label{sec:method}

\begin{figure*}[t]
\centering
\includegraphics[width=0.92\textwidth]{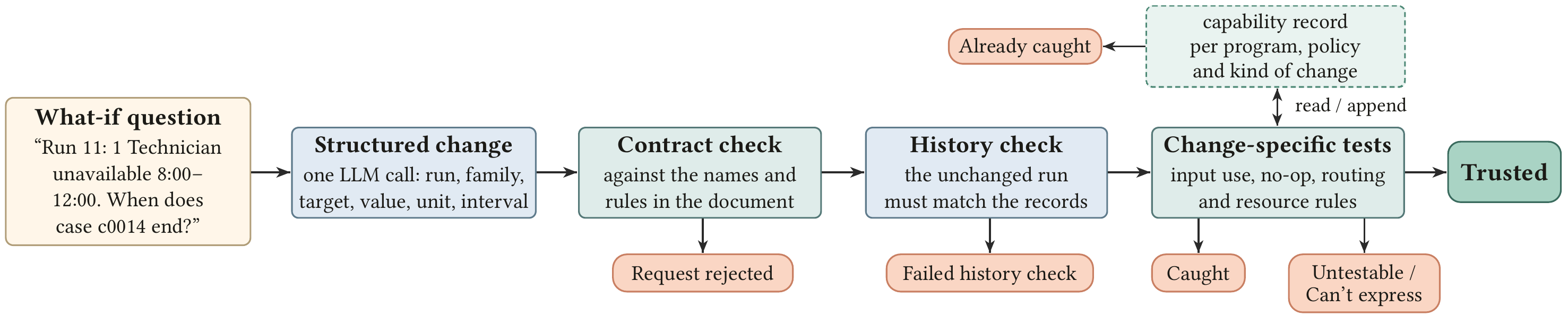}
\caption{How Vouch decides whether to trust a what-if answer, here when case c0014 would finish if one Technician were unavailable. If any step fails, Vouch declines with a reason; it never looks at the final answer itself. Failed tests go into a capability record, so later questions about the same kind of change and target are declined as \Already.}
\Description{A left-to-right flow. A what-if question is turned into a structured change, which passes a contract check, a history check, and tests aimed at the requested change that read and write a capability record. Passing every step gives Trusted. Side exits are labelled Request rejected, Failed history check, Caught, Already caught, Untestable, and Can't express.}
\label{fig:pipeline}
\end{figure*}

Vouch itself is a trust judge, under the definition of \S\ref{sec:problem}. Once fed with a program and a what-if question, it decides whether to release the program's answer. If it decides to decline it, it says what it observed. Instead of examining the answer's value, Vouch asks whether the program handles the requested change on this run. \textbf{Figure~\ref{fig:pipeline}} follows the question about Technician\_7. Firstly, an initial LLM call translates the question into a structured specification, specifically marking one worker in the Technician pool unavailable from 8:00 to 12:00 in Run~11, while a contract check verifies the existence of the target run, pool, and time interval. Next, a history check executes the program against the recorded Run~11 data. Finally, targeted availability tests verify, among other constraints, that no tasks begin on the unavailable worker during the restricted hours.

Vouch is a Python package that connects to any platform through an adapter. The adapter applies a structured change to the program's inputs, runs the program as is, and returns the resulting outputs. We built one adapter per platform, alongside domain-specific parsers to extract names and values from documents. These parsers rely on the fixed layout of our documents. No part of the process relies on a reference simulator, reference answer, or a hidden label. All generated programs run in a network-free sandbox limited to system libraries and the program's own folder.

\subsection{From question to a checked change}
\label{sec:method:request}

One LLM call (Claude Haiku~5.5) turns the question into a structured change: the run, the family, the operation (set a value, add or remove workers, remove a case), the case and target (an activity, a wait, a decision point, a worker pool, a product or a task), the visit, the new value and unit, the time interval, and the outcome to report. We do not rely on this reading. A \emph{contract check} compares it with a description of the process read from the document's tables: the names of runs, cases, activities, pools and decision points, and the stated routes and pool sizes. A request that names an unknown target, matches two targets, uses an operation the family does not allow, or gives an impossible interval or value is declined before any program runs (\Badrequest). The check cannot detect a reading that is well formed but wrong, such as the right pool with the wrong hours; \S\ref{sec:results:main} measures how often that happened.

\subsection{History check}
\label{sec:method:history}

The first test of the program itself is whether it reproduces what happened. Vouch runs the program on the unchanged inputs of the queried run and requires its outcomes to equal the recorded ones exactly: every case end in a workflow, every daily ledger figure in manufacturing, every task event in IT. This is stricter than \otf's own acceptance check, which tolerates timing differences~\cite{otftools2026}. A program that fails is declined on that run (\Failhist).

\subsection{Tests aimed at the requested change}
\label{sec:method:probes}

A history check cannot see a mistake in an input that never varied, so Vouch varies inputs itself. Each test makes a small change, either its own or the one the question asks for, and checks a relation between the program's outputs before and after that any correct simulator must satisfy. The relations follow from the document's rules and records, never from a reference answer; in software-testing terms they are metamorphic relations~\cite{chen2018metamorphic,segura2016survey}, already used to verify and validate simulation models~\cite{raunak2021metamorphic}. We wrote these tests ourselves, one for each kind of change, and the parsers fill in the names and values they need. Vouch chooses the tests by the requested family and target.

The most general test is \emph{input use}. Vouch changes the same kind of input for up to two other cases that the records show were used, for example adding an hour of work. That case's outputs must change, and no output earlier than the moment the changed input is first used may change. This test exposes the loan program, whose outputs do not move when another applicant takes an hour longer. A \emph{no-op} test changes an input for a visit that the records show never happened; nothing may change.

Other families have rules of their own. In the workflow documents the random draw that decided each recorded route is printed (\S\ref{sec:benchmark}), so when routing probabilities change, the stated rule determines which recorded decisions must switch route; if the program's records name decisions, they must show those switches and the new paths of the cases. \otf's free-form records do not name decisions, so there Vouch only checks that the first case that switches changes. Either way, nothing before the first switch may move, and if nothing switches, nothing may change. A worker made unavailable may start no task inside the interval; with added workers, nothing may change before the first task of that pool that had to wait, and that task's case must change. A removed case's records must vanish, and nothing before its arrival may move. Where a program prints records that name tasks, waits and workers, Vouch also checks them against the stated rules for the test runs and for a run of the question's own change: tasks last their stated work, waits their stated delay, workers work only in their hours, and cases follow the paths their printed draws imply. These checks take the program's own sequencing as given, so they cannot produce the queried answer themselves. Manufacturing and IT have analogous checks, which run on the question's own change: nothing before the change may move, every dispatchable order must be dispatched as the policy states, stock and backorders must add up within each day's capacity, and a task's requests must fit the machine it runs on, which has to be up.

A test that cannot run, because no suitable second case exists or the program has no input for the change, counts as neither a pass nor a failure. Vouch trusts an answer only if at least one test that exercises the changed input ran and passed, and none failed; a passing no-op alone is not enough. The input-use and no-op tests only show whether the program reacts to a change, and the rule checks show that it reacts the way the rules say. They do not show that the answer is correct. The tests get a budget of three program runs per question, and the stated-rule checks and some IT tests can add up to two more. The exact selection rules and tolerances are in the artifact.

\subsection{A record of what each program can do}
\label{sec:method:registry}

The same program answers many questions, so Vouch keeps what its tests found in a \emph{capability record}, keyed by the program's code, the adapter version, the policy the program runs under, the family and the operation; each entry also names the target it was tested on. An entry records a failure or a pass on given runs, never that the program is correct. A failure blocks that kind of change for that target wherever the key matches, so a program caught once is not trusted on the same kind of change later (\Already). A pass is reused only on the same run; on new records or under a different policy the tests run again, and passes that depend on the request's own values, such as new route probabilities, are never reused. A policy is identified by a fingerprint of its rule text without the list of runs, so new runs under the same rules share it, while any change to the text, even a harmless rewording, counts as a new policy.

\subsection{Decisions and variants}
\label{sec:method:terms}

Every question resolves to one of seven named outcomes. The first is the \textbf{\Trusted} outcome, the only outcome that \emph{releases} the answer.
\begin{rulebox}
\small
\noindent\textbf{When Vouch trusts an answer.} For a question $q$ about run $r$, answered by program $P$:
\[\begin{aligned}
\mathit{Trust}(P,q) \iff{} & C(q) \wedge H(P,r) \wedge \neg F(P,k)\\
& \wedge\ \forall t\in T:\ o(t)\neq \mathsf{fail}\\
& \wedge\ \exists t\in T\setminus N:\ o(t)= \mathsf{pass}
\end{aligned}\]
\footnotesize
where $C(q)$: the structured change passes the contract check (\S\ref{sec:method:request}); $H(P,r)$: $P$ reproduces the recorded history of $r$ (\S\ref{sec:method:history}); $F(P,k)$: the record holds a failure of $P$ for this kind of change and target, $k$ (\S\ref{sec:method:registry}); $T$: the tests for $q$, run now or reused from the same run, each with outcome $o(t)$ of pass, fail or not run; $N\subseteq T$: the no-op tests (\S\ref{sec:method:probes}). In manufacturing and IT, the last condition must hold for the question's own change and for a change of the same kind elsewhere.
\end{rulebox}
\noindent \textbf{Request rejected} halts invalid queries at the contract check. The remaining five outcomes are  either declines or abstentions. Declines' outcomes flag detected errors: \textbf{Caught} (a test failed on the current question), \textbf{Already caught} (a prior recorded failure for that change type and target), and \textbf{Failed history check} (the program fails to reproduce the baseline run). Abstentions' outcomes return no evidence:  \textbf{Untestable} (when no test can be executed to gather evidence) and \textbf{Can't express} (the platform adapter cannot map the requested change to any program input).

To measure which components matter, we evaluate four ablated baselines on the same code: (1) the contract check alone; (2) the history-check judge (contract plus history checks); (3) generic tests, which add input-use tests on randomly sampled families; and (4) program-wide tests, which run one input-use test per family to assign a single shared verdict to all questions on a given program and run. Generic tests lack routing, worker and stated-rule checks, so their difference from Vouch measures the whole choice of tests.

\section{Benchmark}
\label{sec:benchmark}

\begin{table*}[t]
\centering
\caption{Domains, change families and test questions. The no-effect column counts the test questions whose change, though valid, leaves the queried outcome unchanged.}
\label{tab:benchmark}
\footnotesize
\begin{tabular}{@{}l p{4.6cm} p{6.6cm} r r@{}}
\toprule
Domain & Source of inputs and rules & Change families (what a question changes) & Test qs. & No-effect \\
\midrule
Loan application & Published process model and simulation parameters~\cite{chapela2023zenodo}; 17 activities, loops, 6 worker pools & activity duration; timer delay; the route one case takes; routing probabilities; worker availability (workers added or unavailable); case removal & 120 & 24 \\
Pharmacy & Same source; 12 activities, loops, 3 worker pools & the same six families & 120 & 24 \\
Manufacturing & SupplyGraph production and orders~\cite{wasi2024supplygraph}; written dispatch policy~\cite{otftools2026} & order removal; production quantity; production timing; dispatch capacity & 96 & 16 \\
IT operations & Google 2011 cluster trace~\cite{reiss2011google}; written priority scheduler on 8 machines~\cite{otftools2026} & job removal; machine-failure removal; task priority; task resource request; machine removal & 120 & 20 \\
\midrule
Policy variants & Loan and Pharmacy with one more worker in one pool (P) or with the rules reworded (H) & the six workflow families & 91 & 22 \\
\bottomrule
\end{tabular}
\end{table*}

Evaluating whether a released answer is accurate requires what-if questions with known ground-truth answers, posed to programs generated without seeing those questions. This section describes our evaluation domains, our methodology for guaranteeing and verifying exact ground-truth answers, and the candidate programs we assess.

\paragraph{Domains.} Table~\ref{tab:benchmark} lists the four domains. The two workflows come from an open repository of parameterized business-process models~\cite{chapela2023zenodo,chapela2024extraneous}; Manufacturing and IT operations reuse the documents of a prior benchmark of process records~\cite{otftools2026}, in which real production, order and cluster-trace logs run under written dispatch and schedule policies. A workflow run contains 15 cases, a manufacturing run covers one plant over 14 days, and an IT operations run captures one window of cluster activity.

\paragraph{Exact answers.} A what-if question has an exact answer only if the random parts of a run are pinned down, and a shared random seed does not do this. In a simulator that draws from one stream in execution order, such as Prosimos~\cite{lopezpintado2023prosimos}, adding one case shifts the draws of every later case, so the ``what if'' compares two unrelated runs. We therefore wrote the workflow semantics down first and built a deterministic reference engine for them. Every random quantity is tied to a stable identity (case, activity or decision, and visit) and printed in the document: planned work, planned waits, and the draw that decided each route. Removing a case leaves every other case's draws unchanged, and a new routing probability is applied to the same draw. This is the common-random-numbers technique~\cite{law2015simulation}: the question queries the same realized run under one change, and the document contains everything needed for an exact answer. Manufacturing and IT are deterministic by construction.

\paragraph{Checked reference answers.} Reference answers are computed by deterministic reference code, which was itself checked. For the workflows, a separate checker was written from the semantics specification alone, without the engine's code or outputs; engine and checker agree on 18 of 18 hand-worked cases, on 566 fixed and random scenarios in every output field, and on all 240 workflow test questions. Independent LLM agents given only the document text also matched 24 of 24 sampled answers. For manufacturing and IT, the original simulators and a second implementation fed only by values parsed back from the documents agree on all 216 test questions, and blind samples matched 16 of 16 (by hand) and 20 of 20 (by agents writing their own schedulers).

\paragraph{Questions and splits.} Each question is written from a fixed template for its family and selected by a fixed rule, without regard to how any system answers it. Data are split by run: workflow runs 1--4 are for development and runs 5--12 for testing, and IT windows that share jobs, or manufacturing periods that share order days, stay on the same side. The development questions (112 workflow, 108 manufacturing and IT) were used to design Vouch and are not reported as evidence.

\paragraph{Policy variants.} Programs are meant to be reused after the rules change, so the policy set asks questions under rules the program was not built for. In variant P (a policy change) one pool gets one more worker (Loan appraisers 2$\to$3; Pharmacy pharmacists 1$\to$2), stated in the document. In variant H (a harmless rewording) the parameters are unchanged and 24 sentences of the rules are reworded; four independent LLM agents reproduced every H run from the H text alone. A program that hard-codes the old pool size should fail on P, and nothing should change on H.

\paragraph{Platforms and programs.} Three platforms build programs. \otf~\cite{otftools2026} (first platform) compiles a document into functions that read its records and a simulator that runs them, and accepts a program if it reproduces hidden recorded outcomes closely enough. Its original code often failed on our documents for mechanical reasons, such as reading records from the wrong place, so we use a revised generation path, which we call R1, that makes record locations correct by construction and retries reading; R1 gives no hints about process logic and keeps the acceptance check. The unchanged original \otf accepted 32 of 60 builds; its programs answer only through \otf's own path and are reported in the artifact. \devs~\cite{chen2026devs} (second platform) generates discrete-event models in stages from a specification, to which we add one generic input format for records, plans and probabilities. The coding agent (third platform) is a single model asked to write a simulator for the same specification and run it once on a published example input. Claude Sonnet~5.5 and Haiku~5.5 build on every platform, and three open-weight models, gpt-oss-120b~\cite{openai2025gptoss}, Gemma~4 31B and Qwen3.5~9B, build on \otf only. Of 101 build attempts, 86 were accepted: 48 on \otf R1, where Qwen produced only 3 of 12, 22 on \devs and all 16 from the coding agent. Builds are independent and none was selected by its results. Builders had no access to test documents, reference answers or reference code, which we checked with sandbox isolation tests and by scanning every program for test-only names and values. The main test set comprises 9{,}792 program--question pairs across these 86 programs, while the policy set contains 1{,}411 pairs drawn from the 31 Claude-built workflow programs.

\section{Trust Judges Compared}
\label{sec:judges}

Every judge decides about each of the 9{,}792 program--question pairs, whose answers are fixed. All judges receive the same structured change (\S\ref{sec:method:request}), so differences come from how they judge, not from how the question was read. No judge sees reference answers, reference simulators, the hidden correct reading of the question, other programs, other judges' verdicts or build histories.

\paragraph{Rule-based judges.} Four judges run on every pair without model calls. The \emph{platform check} trusts every answer from an accepted program; it is the single verdict a platform gives today, and for the coding agent it means only that the program ran. The \emph{history check} trusts an answer if the program reproduces the queried run exactly. \emph{Generic tests} are the reduced version of Vouch in \S\ref{sec:method:terms}. \emph{Disagreement} trusts an answer if it agrees with a \emph{reading answer}, which Claude Haiku gives after reading the rules and the queried run, without any program.

\paragraph{LLM judges.} Each LLM judge returns a verdict, a confidence from 0 to 100 that the answer is exactly right, and a one-line reason; a reply that cannot be parsed, is cut off, is refused or runs out of turns counts as a decline. The \emph{answer-level judge}~\cite{zheng2023judging} makes one call that sees the rules, the queried run's records, the program's code, the question, the structured change, the answer and the unchanged answer. The \emph{program-level judge} makes one call per program and run, without the question and answer, and its verdict covers every question on that run. The \emph{ReAct judge}~\cite{yao2023react} alternates reasoning with actions; its one tool runs Python in a sandbox holding the program and the same document view, for at most eight turns. It is told what to return and nothing about how to test, and since it can run the program on any input, it has at least Vouch's access to execution. The answer-level and ReAct judges also see the question text and the answer, which Vouch never uses.

\paragraph{A decision model as judge.} After the main run we added Jev (jev-1.14.0), a hosted model from TypeSafe~\cite{typesafe2026jev} that writes no text and returns the probability that the answer is exactly right. It receives the answer-level judge's input without the reply-format lines, trusts an answer when that probability is at least 0.5, and cannot run code; what is known about its accuracy comes from the vendor. Inputs the service rejected as too long for its default limit were sent again once in its long-context mode, a rule that depends only on length (\rJevLongRows{} of 3{,}000 pairs, almost all from the loan domain).

\paragraph{Models and subsets.} Interpretation and reading answers use Claude Haiku~5.5 for every platform. LLM judging is costly, so it ran on fixed subsets of answered pairs, drawn by pair identifier with a fixed seed before any judging, balanced across domain, platform, builder model and family, and taken from the main and policy sets together. The Haiku answer-level judge and Jev saw the same 3{,}000 pairs, the Haiku ReAct judge 1{,}000 of them, and the Haiku program-level judge every program and run. A shared subset of 150 pairs was judged by all five models using the answer-level judge, and by four models (Haiku, gpt-oss, Gemma and Qwen) using the ReAct judge; Sonnet's ReAct judge evaluated 50 of these pairs. For each judge, we report results on the Claude-built main-test pairs within its evaluated subset (for example, 2{,}028 of the 3{,}000 pairs).

\paragraph{Analysis plan.} Vouch, the runner, the scorer, the judge prompts, the questions, the splits and the program list were frozen and hashed before any test question was run. Later changes were raised spending limits, the open-weight and Sonnet ReAct judges added on the shared subset, the Jev judge added after the main run, subsets drawn from the policy set as well, the run-level analysis described below, and a rerun without the capability record (\S\ref{sec:results:families}); no frozen code, prompt or program changed, and every change is logged. Uncertainty comes from a two-level bootstrap with 2{,}000 resamples of programs and then questions, with the coverage match redone in every resample and every difference computed on the same resamples. Due to the fact that questions on one run share records and draws, we repeated the analysis resampling whole runs, and call a difference significant only if its 95\% interval excludes zero under both versions.

\section{Results}
\label{sec:results}

\subsection{Errors that survive the history check}
\label{sec:results:main}

\begin{table*}[t]
\centering
\caption{Rule-based judges and Vouch on the main test set. Cov.: share of all program--question pairs whose answer the judge releases (\%). Wrong: released wrong answers (for random refusal, the expected number). $\Delta$K: correct answers kept by Vouch minus those kept by the judge, in percentage points of all pairs; negative values mean the judge keeps more correct answers. Bold: the 95\% interval of $\Delta$K excludes zero under both bootstrap analyses (\S\ref{sec:judges}). The contract check alone releases exactly what the platform check releases.}
\label{tab:main}
\footnotesize
\setlength{\tabcolsep}{2.6pt}
\begin{tabular}{@{}l rrr rrr rrr rrr rrr@{}}
\toprule
& \multicolumn{12}{c}{Built by Sonnet and Haiku} & \multicolumn{3}{c}{Built by open-weight models} \\
\cmidrule(lr){2-13}\cmidrule(l){14-16}
& \multicolumn{3}{c}{\otf R1 (24 progs)} & \multicolumn{3}{c}{\devs (22)} & \multicolumn{3}{c}{Coding agent (16)} & \multicolumn{3}{c}{All three (62)} & \multicolumn{3}{c}{\otf R1 (24)} \\
& \multicolumn{3}{c}{2{,}736 pairs} & \multicolumn{3}{c}{2{,}496} & \multicolumn{3}{c}{1{,}824} & \multicolumn{3}{c}{7{,}056} & \multicolumn{3}{c}{2{,}736} \\
\cmidrule(lr){2-4}\cmidrule(lr){5-7}\cmidrule(lr){8-10}\cmidrule(lr){11-13}\cmidrule(l){14-16}
Judge & Cov. & Wrong & $\Delta$K & Cov. & Wrong & $\Delta$K & Cov. & Wrong & $\Delta$K & Cov. & Wrong & $\Delta$K & Cov. & Wrong & $\Delta$K \\
\midrule
Platform check only & 87.0 & 243 & \sig{$-7.6$} & 87.8 & 34 & \sig{$-8.4$} & 97.6 & 0 & $-0.7$ & 90.0 & 277 & \sig{$-6.1$} & 61.0 & 1161 & \sig{$-14.4$} \\
History check & 82.4 & 202 & \sig{$-4.5$} & 79.0 & 5 & \sig{$-0.7$} & 97.6 & 0 & $-0.7$ & 85.1 & 207 & \sig{$-2.2$} & 4.9 & 18 & $-0.1$ \\
Generic tests & 82.1 & 200 & \sig{$-4.3$} & 78.7 & 5 & $-0.5$ & 97.6 & 0 & $-0.7$ & 84.9 & 205 & \sig{$-2.0$} & 4.9 & 18 & $-0.1$ \\
Disagreement & 59.7 & 29 & \sig{$+11.8$} & 68.3 & 4 & $+9.9$ & 75.0 & 0 & \sig{$+21.9$} & 66.7 & 33 & \sig{$+13.8$} & 15.5 & 18 & \sig{$-10.6$} \\
Random refusal & 70.5 & (197) & \sig{$+7.2$} & 78.2 & (30) & $+1.0$ & 96.9 & (0) & 0.0 & 80.1 & (246) & \sig{$+3.4$} & 4.2 & (81) & $+2.9$ \\
\midrule
Program-wide tests & 82.1 & 202 & \sig{$-4.2$} & 78.7 & 4 & $-0.5$ & 97.6 & 0 & $-0.7$ & 84.9 & 206 & \sig{$-2.0$} & 4.9 & 18 & $-0.1$ \\
Vouch & 70.5 & 0 & -- & 78.2 & 5 & -- & 96.9 & 0 & -- & 80.1 & 5 & -- & 4.2 & 2 & -- \\
\bottomrule
\end{tabular}
\end{table*}

\begin{table*}[t]
\centering
\caption{LLM judges and a decision model (Jev) against Vouch on Claude-built programs (main test set). Each judge ran on its own fixed subset of $n$ answered pairs, so coverage and error here are shares of answered pairs. Err.: error among released answers (\%), with the number of wrong answers in parentheses (expected number when matched). Matched: the judge's most confident answers, up to Vouch's count in each domain, platform and builder model. Own verdict: the judge's own decisions. $\Delta$: Vouch minus the judge, in points, with 95\% interval; positive $\Delta$ precision favors Vouch, negative $\Delta$ kept means the judge keeps more correct answers. Bold: interval excludes zero under both bootstrap analyses; $^\ast$excludes zero only when resampling questions (resampling runs: [$-13.5$, 0.5]). Failed: replies counted as declines (\S\ref{sec:judges}). Jev's own verdicts depend on its 0.5 threshold.}
\label{tab:llm}
\footnotesize
\setlength{\tabcolsep}{3.2pt}
\begin{tabular}{@{}l r r r r r r r r@{}}
\toprule
& & \gate & \multicolumn{2}{c}{Judge, matched} & \multicolumn{3}{c}{Judge, own verdict} & \\
\cmidrule(lr){3-3}\cmidrule(lr){4-5}\cmidrule(lr){6-8}
Judge & $n$ & Err. & Err. & $\Delta$ precision & Cov. & Err. & $\Delta$ kept & Failed replies \\
\midrule
\multicolumn{9}{@{}l}{\textit{Answer-level judge}} \\
\quad Haiku & 2{,}028 & 0.2 (3) & 1.1 (20) & \sig{$+0.9$ [0.2, 2.0]} & 85.7 & 0.6 (10) & $+3.1$ [$-2.7$, 7.8] & 1/3{,}000 \\
\quad Sonnet & 101 & 0.0 (0) & 0.0 (0) & 0.0 [0.0, 0.0] & 94.1 & 0.0 (0) & $-9.9$ [$-23.4$, 1.1] & 0/150 \\
\quad gpt-oss & 101 & 0.0 (0) & 3.2 (3) & $+3.2$ [0.0, 6.2] & 60.4 & 3.3 (2) & \sig{$+25.7$ [6.2, 43.6]} & 0/150 \\
\quad Gemma & 101 & 0.0 (0) & 2.7 (2) & $+2.7$ [0.0, 5.5] & 81.2 & 1.2 (1) & $+4.0$ [$-10.7$, 18.4] & 0/150 \\
\quad Qwen & 101 & 0.0 (0) & 0.6 (1) & $+0.6$ [0.0, 1.8] & 80.2 & 1.2 (1) & $+5.0$ [$-12.3$, 22.6] & 5/150 \\
\multicolumn{9}{@{}l}{\textit{Program-level judge}} \\
\quad Haiku & 6{,}352 & 0.1 (5) & 3.5 (196) & \sig{$+3.4$ [1.7, 5.6]} & 76.4 & 3.3 (160) & \sig{$+14.9$ [9.6, 21.0]} & 0/1{,}778 \\
\multicolumn{9}{@{}l}{\textit{ReAct judge}} \\
\quad Haiku & 679 & 0.0 (0) & 1.4 (8) & \sig{$+1.4$ [0.4, 3.9]} & 93.1 & 1.3 (8) & $-5.6$ [$-11.7$, $-0.5$]$^\ast$ & 3/1{,}000 \\
\quad Sonnet & 32 & 0.0 (0) & 0.0 (0) & 0.0 [0.0, 0.0] & 87.5 & 0.0 (0) & $-9.4$ [$-30.3$, 0.0] & 1/50 \\
\quad gpt-oss & 101 & 0.0 (0) & 2.3 (2) & $+2.4$ [0.0, 5.5] & 53.5 & 1.8 (1) & \sig{$+31.7$ [11.1, 50.0]} & 38/150 \\
\quad Gemma & 101 & 0.0 (0) & 2.5 (2) & $+2.5$ [0.0, 5.0] & 90.1 & 1.1 (1) & $-5.0$ [$-20.2$, 9.9] & 3/150 \\
\quad Qwen & 101 & 0.0 (0) & 2.3 (2) & $+2.4$ [0.0, 5.7] & 71.3 & 0.0 (0) & $+12.9$ [$-7.5$, 32.6] & 24/150 \\
\multicolumn{9}{@{}l}{\textit{Decision model, answer-level input}} \\
\quad Jev & 2{,}028 & 0.2 (3) & 2.3 (42) & \sig{$+2.2$ [0.8, 4.2]} & 73.5 & 3.1 (46) & \sig{$+17.0$ [10.8, 23.6]} & 0/3{,}000 \\
\bottomrule
\end{tabular}
\end{table*}

\begin{figure}[t]
\centering
\begin{tikzpicture}
\begin{axis}[
  width=\columnwidth, height=3.4cm,
  xmin=0, xmax=104, ymin=-0.6, ymax=12,
  xlabel={Coverage of answered pairs (\%)}, ylabel={Error among trusted (\%)},
  xlabel style={font=\footnotesize, yshift=3pt}, ylabel style={font=\footnotesize, yshift=-4pt},
  tick label style={font=\footnotesize},
  xtick={0,20,40,60,80,100}, ytick={0,2,4,6,8,10,12},
  legend style={font=\scriptsize, at={(0.5,-0.31)}, anchor=north, draw=none, column sep=4pt},
  legend columns=3, legend cell align=left, clip=true,
]
\addplot[blue!70!black, thick, mark=none] coordinates {(0.39,33.33) (1.42,18.18) (6.84,9.43) (14.71,6.14) (17.16,5.26) (30.84,3.77) (40.00,2.90) (40.26,2.88) (50.19,3.08) (57.16,2.71) (60.26,2.57) (62.45,2.48) (62.97,2.46) (66.06,2.34) (67.23,2.30) (68.00,2.28) (68.26,2.27) (68.52,2.26) (68.90,2.25) (70.06,2.21) (74.71,2.07) (74.84,2.07) (81.29,1.90) (81.81,1.89) (82.06,1.89) (82.19,1.88) (83.87,1.85) (84.77,1.83) (85.55,1.81) (86.71,1.79) (86.84,1.78) (87.48,1.77) (88.52,2.04) (90.19,2.86) (90.32,2.86) (92.39,4.19) (92.52,4.18) (92.90,4.44) (97.68,8.06) (99.74,9.96) (100.00,10.19)};
\addlegendentry{Answer ($n{=}775$)}
\addplot[orange!85!black, thick, dashed, mark=none] coordinates {(1.15,66.67) (2.68,28.57) (8.81,13.04) (11.88,12.90) (20.69,9.26) (35.25,5.43) (45.59,5.88) (51.72,5.19) (57.85,5.30) (59.39,5.16) (59.77,5.13) (60.15,5.10) (65.90,4.65) (72.41,4.23) (75.86,4.04) (77.01,3.98) (79.31,3.86) (80.08,3.83) (81.99,3.74) (85.44,3.59) (86.59,3.54) (88.89,3.88) (90.04,5.11) (91.19,5.46) (93.49,7.38) (93.87,7.35) (96.55,9.52) (99.23,10.81) (100.00,10.73)};
\addlegendentry{ReAct ($n{=}261$)}
\addplot[black!55, thick, dotted, mark=none] coordinates {(0.46,36.36) (2.02,18.75) (2.31,16.36) (5.29,11.11) (6.18,9.52) (6.60,8.92) (6.81,8.64) (8.99,6.54) (25.42,9.75) (28.87,9.17) (32.69,10.15) (38.36,10.73) (39.71,10.90) (64.03,10.30) (68.07,9.94) (68.82,10.07) (69.58,9.96) (70.13,9.89) (70.17,9.88) (72.90,9.74) (76.97,9.50) (84.45,9.00) (85.71,8.97) (86.39,8.95) (87.06,9.03) (93.53,9.61) (97.10,9.95) (99.29,10.07) (100.00,10.21)};
\addlegendentry{Program-level ($n{=}2{,}380$)}
\addplot[only marks, forget plot, mark=square*, mark size=1.8pt, black!75] coordinates {(100.00,10.21)};
\node[font=\scriptsize, left, xshift=-1pt, yshift=3pt] at (axis cs:100.00,10.21) {no check};
\addplot[only marks, forget plot, mark=triangle*, mark size=2.4pt, black!75] coordinates {(94.71,8.96)};
\node[font=\scriptsize, left, xshift=-1pt, yshift=-3pt] at (axis cs:94.71,8.96) {history};
\addplot[only marks, forget plot, mark=diamond*, mark size=2.4pt, black!75] coordinates {(68.66,1.77)};
\node[font=\scriptsize, left] at (axis cs:68.66,1.77) {disagreement};
\addplot[only marks, forget plot, mark=*, mark size=2.6pt, red!75!black] coordinates {(81.05,0.00)};
\node[font=\scriptsize, above, yshift=1pt] at (axis cs:81.05,0.00) {\gate};
\end{axis}
\end{tikzpicture}
\caption{Error against coverage on \otf R1 programs built by Sonnet and Haiku, where 243 of the 277 wrong answers occur. Curves sweep each Haiku judge's confidence threshold on the pairs it judged ($n$); markers are judges with a single operating point, on all 2{,}380 answered pairs. \gate trusts 81\% of these pairs with no wrong answer, at a lower coverage than the history check (95\%).}
\Description{A line chart with coverage on the x axis and error on the y axis. Three dashed or solid curves for Haiku judges stay between about 2 and 10 percent error at coverages above 20 percent. A red dot for Vouch sits at about 81 percent coverage and 0 percent error. Points for the history check and no added check sit near 95 and 100 percent coverage at about 9 and 10 percent error; disagreement sits at about 69 percent coverage and 2 percent error.}
\label{fig:riskcov}
\end{figure}
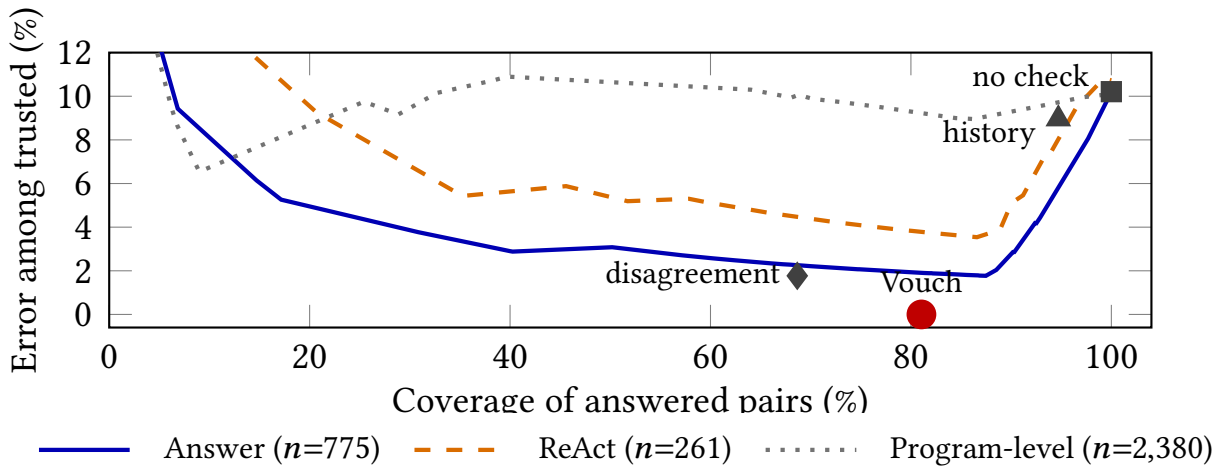

Table~\ref{tab:main} shows what each rule-based judge releases. With only the platform check, Claude-built programs answer \rNoCheckCovClaude\% of the pairs, and \rNoCheckWrongClaude{} of those answers are wrong (\rNoCheckErrClaude\%). The history check leaves \rHistWrongClaude{} of them, and Vouch leaves \rGateWrongClaude, an error of \rGateErrClaude\% at \rGateCovClaude\% coverage. A random refusal at Vouch's coverage keeps the \rRandomErrClaude\% error and 3.4 points fewer correct answers, so declining the same number of answers at random would not produce Vouch's gain.

Most of these errors come from one platform. On \otf R1 programs, \rOtfHistWrong{} wrong answers pass the history check and Vouch releases none of them. Of all \rNoCheckWrongClaude{} wrong answers, \rRouteProbWrong{} follow a change of routing probabilities, which the history check cannot exercise because the recorded probabilities never vary. On \devs the history check already leaves only 5 wrong answers, and so does Vouch. The coding agent's programs, and all programs in manufacturing, give no wrong answer at all. Programs written by open-weight models mostly fail the history check: with \rNoCheckErrOllama\% error under the platform check, Vouch releases only \rGateCovOllama\% of their pairs, and a disagreement check keeps 10.6 points more correct answers.

These errors are not misreadings of the question: for every one of the \rNoCheckWrongClaude{} wrong answers, the structured change equalled the hidden correct reading. Interpretation was exact on \rInterpWf{} workflow and \rInterpMl{} manufacturing and IT questions; the seven misreadings cost only coverage (\rMlCovInterp\% instead of \rMlCovOracle\% with the correct reading). The faults lie in the programs, and in many of them: \rProgsWithWrong{} of the \rProgsClaude{} Claude-built programs give a wrong answer. Vouch removes every wrong answer in 15 of them. The \rProgsWrongKept{} that keep one or two are \devs programs, and no \otf R1 program contributes more than \rOtfLargestProgWrong{} of the 243.

Vouch pays for this in correct answers. It declines \rRightLostCountClaude{} of 6{,}075 correct answers (\rRightLostClaude\%) and keeps 2.2 points fewer correct answers than the history check, 4.5 on \otf R1. The loss is largest in IT, where the history check keeps 5.4 points more with the same three wrong answers: once a program fails a test of removing a particular machine, Vouch declines every later question about that machine, and \rItAlreadyRight{} of those answers were correct (\S\ref{sec:results:families}). Generic and program-wide tests release almost exactly what the history check releases (205 and 206 wrong), so random input-use tests do not find these errors; rather, Vouch's routing and worker tests do.

\subsection{Detection or abstention}
\label{sec:results:families}

\begin{table}[t]
\centering
\caption{Why Vouch declined, by change family (Claude-built programs, main test set). Wrong: wrong answers under the platform check alone. Kept: wrong answers Vouch releases. Detect and Abstain: wrong answers removed / correct answers lost by declines that detect a problem (\Caught, \Already, \Failhist) and by declines that abstain (\Untestable, \Cantexpress). Manufacturing is omitted: no program gave a wrong answer there, and Vouch abstained on 9 correct answers. The totals include it.}
\label{tab:families}
\footnotesize
\setlength{\tabcolsep}{3pt}
\begin{tabular}{@{}l r r r c c@{}}
\toprule
Family & Pairs & Wrong & Kept & Detect & Abstain \\
\midrule
\multicolumn{6}{@{}l}{\textit{Workflows (Loan and Pharmacy)}} \\
Activity duration & 620 & 11 & 0 & 11/55 & 0/0 \\
Timer delay & 620 & 15 & 0 & 15/67 & 0/0 \\
Route draw (one case) & 620 & 5 & 0 & 5/44 & 0/14 \\
Routing probabilities & 620 & 195 & 0 & 195/70 & 0/0 \\
Worker availability & 620 & 33 & 1 & 32/36 & 0/0 \\
Case removal & 620 & 11 & 1 & 10/20 & 0/0 \\
\multicolumn{6}{@{}l}{\textit{IT operations}} \\
Job removal & 360 & 3 & 1 & 2/1 & 0/0 \\
Failure removal & 360 & 1 & 1 & 0/4 & 0/14 \\
Task priority & 360 & 0 & 0 & 0/5 & 0/0 \\
Task resource request & 360 & 2 & 0 & 2/6 & 0/6 \\
Machine removal & 360 & 1 & 1 & 0/79 & 0/0 \\
\midrule
All families & 7{,}056 & 277 & 5 & 272/387 & 0/43 \\
No-effect controls & 1{,}300 & 16 & 0 & 16/90 & 0/37 \\
\bottomrule
\end{tabular}
\end{table}

A judge can lower its error by declining whatever it cannot check, so we ask why Vouch declined each wrong answer (Table~\ref{tab:families}). Of the 272 it removed, \rCaughtWrong{} were \Caught by a test on the question itself, \rAlreadyWrong{} were \Already because the program had failed the same test earlier, and 70 came from Failed history check. None was removed by abstaining. Of the \rUntestableRight{} correct answers declined as \Untestable, \rUntestableRightNoEffect{} are no-effect controls on which no test of the change could run.

An example from the routing family shows detection at work. In Run~05 of the loan process, a question asks when case c0005 would end if decision D2 sent 30\% of cases one way and 70\% the other. All six Claude-built \otf R1 loan programs pass the history check and answer 47{,}253 seconds, the recorded value; the correct answer is 207{,}014. Vouch's routing test looks at the earliest decision on that run that the new probabilities must switch: the draw printed for another case, c0004, at D2 now requires it to take the other route, but none of the programs changes c0004's records, and Vouch declines (\Caught). The programs ignore the stated probabilities, which never vary in the recorded history.

Detection also costs correct answers. Of the 430 declined, 277 come from programs that fail the history check, 43 from abstention and 110 from failed tests, which can fail on a case or input than the one the question touches. Retesting every question instead of consulting the record catches the same wrong answers and releases \rFreshRightRecovered{} more correct ones, at \rFreshExecPerQ{} instead of 1.54 program runs per question, so the record saves runs, not errors.

\subsection{Comparison with LLM judges}
\label{sec:results:llm}

Table~\ref{tab:llm} compares Vouch with LLM judges, the usual alternative to testing, each on its own subset, and Figure~\ref{fig:riskcov} shows the whole trade-off on \otf R1, where the errors are. At equal coverage the three Haiku judges accept more wrong answers than Vouch: the answer-level judge \rAnsJudgeErrClaude\% against \rGateErrAnsSub\% on the same \rAnsJudgeN{} pairs, the program-level judge \rProgJudgeErrClaude\% against 0.1\%, and the ReAct judge \rReactErrClaude\% against \rGateErrReactSub\%; all three differences are significant. No threshold closes the gap: above 5\% coverage no Haiku judge gets below 1.8\% error on \otf R1, while Vouch releases 81\% of those answers with none wrong.

Sonnet and the open-weight judges ran on 32 to 101 Claude-built pairs containing at most four wrong answers, and at equal coverage Vouch ties with all of them. On its own verdicts, Sonnet's answer-level judge trusts \rSonnetAnsOwnCov\% of these pairs with no wrong answer, against Vouch's 84.2\%; Sonnet and the Haiku ReAct judge tend to keep more correct answers than Vouch, though neither difference holds when whole runs are resampled. Being able to run the program did not make the ReAct judge a tester: Haiku's ran it on only \rReactRanProgHaiku\% of its pairs and Sonnet's on 11 of 50, deciding the rest from reading alone.

Jev judged the same \rJevN{} pairs as the Haiku answer-level judge but ranks answers less well (AUROC \rJevAUROC{} against \rHaikuAnsAUROC{}). At Vouch's coverage it accepts \rJevErrClaude\% wrong answers against \rGateErrAnsSub\%, and \rJevErrOtf\% on \otf R1, where Vouch accepts none; its long-context mode does not explain this, since 43 of the 46 wrong answers it trusted on its own verdicts came from inputs sent in the default mode. On the 123 Claude-built pairs of the shared 150 it ties with Vouch. For programs written by open-weight models, its own verdicts keep more correct answers than Vouch ($\Delta$ kept \rJevDKeptOllama{} points), as Haiku's do, with 42 wrong answers against Vouch's 2.

\subsection{When the rules change}
\label{sec:results:reuse}

Programs are meant to be reused after the rules change. Under P, where one pool has one more worker, the programs answer \rPolPNoCheckCov\% of the questions under the platform check alone, and \rPolPNoCheckErr\% of those answers are wrong. Every program fails the history check on every P run, so Vouch releases nothing, while a disagreement check keeps \rPolPDisCov\% of the questions without a wrong answer. Under H, where the rules are reworded but mean the same, Vouch treats the text as a new policy, tests again (1.45 extra runs per question), and releases \rPolHGateCov\% of the answers with none wrong; the history check alone releases \rPolHHistCov\% with 35 wrong. No P or H document shares a fingerprint with a test document, so nothing recorded under the original rules was carried over, and this set does not test whether recorded results transfer correctly across policies.

\subsection{Cost}
\label{sec:results:cost}

Vouch makes no model calls beyond the interpretation every judge shares; its cost is program runs. It used 1.54 extra runs per question on average (median \rGateExecMedian), fewer than generic tests (2.62) or program-wide tests (3.21) because results are reused within a run (\rFreshExecPerQ{} without the record), and all rule-based judges together took a median of \rGateSecPerQ{} seconds per question. The Haiku answer-level judge cost \$\rAnsJudgeUSDPerRow{} and \rAnsJudgeSecPerRow{} seconds per pair, the Haiku ReAct judge \$\rReactUSDPerRow{} and \rReactSecPerRow{} seconds, and Jev \$\rJevUSDPerRow{} and \rJevSecPerRow{} seconds. The whole study, including development and failed runs, cost about \$\rTotalUSD{} in model calls.

\section{Related Work}
\label{sec:related}

\paragraph{LLM-generated simulators.} Several systems build executable models from text. \otf compiles process-record documents into reusable readers and simulators and accepts a program if it reproduces hidden recorded outcomes~\cite{otftools2026}. \devs generates discrete-event models from specifications and checks their event traces, including on intervention inputs, against temporal, causal and semantic constraints derived from the specification~\cite{chen2026devs}. G-Sim proposes a simulator's structure with an LLM and calibrates it against data~\cite{holt2025gsim}, and program-aided reasoning has a model write code and trusts its execution~\cite{gao2023pal}. A validation-aware framework for LLM-driven simulation separates whether a simulation runs from whether its numbers are valid~\cite{yoo2026validation}, and work on LLM agents that supply Modelica models asks what an acceptance test certifies~\cite{frochte2026modelica}. We study a different decision: whether to release one answer, for one requested change, from a program already accepted, measured against independently checked answers.

\paragraph{Validity for an intended use.} Simulation practice holds that a model is valid only for the purpose it was built for, and that comparison with recorded history is one validation technique among several~\cite{sargent2013vv}. Machine learning reached a related conclusion under the name underspecification: models that score equally well on held-out data can behave very differently on inputs that data does not exercise~\cite{damour2022underspecification}. We apply the principle per question, when an answer is about to be released, and measure when doing so changes the outcome.

\paragraph{Testing simulation programs.} Vouch's tests are metamorphic relations, properties that must hold between runs on related inputs when no oracle exists~\cite{chen2018metamorphic,segura2016survey}, which have been used to verify and validate simulation models~\cite{raunak2021metamorphic}. Fixing random draws so that a factual and a changed run can be compared exactly is the common-random-numbers technique~\cite{law2015simulation}, the same device that holds exogenous variables fixed in structural counterfactuals~\cite{pearl2009causality}.

\paragraph{Structured requests and abstention.} Turning a what-if request into a checkable specification has been used to build interactive what-if interfaces, where many specifications are wrong even when they compile~\cite{gathani2026praxa}, and a deterministic kernel that answers only bounded question shapes keeps a generative front end from choosing the values it returns~\cite{wu2026never}. Our contract check plays a similar role, but the decision to release rests on tests of an untrusted program. From selective prediction~\cite{elyaniv2010selective,geifman2017selective} and its extension to agent trajectories~\cite{opoku2026toolchain} we adopt comparison at equal coverage; Vouch makes a yes-or-no decision without calibration.

\paragraph{Models as judges.} LLMs are widely used to grade other models' outputs~\cite{zheng2023judging} and can favor outputs of their own family~\cite{panickssery2024self}, which matters here because the same families build and judge programs.

\section{Limitations}
\label{sec:limits}

\paragraph{The reference and the builders share a model family.} Claude models helped write the reference engine. A separately started Claude model wrote the independent checker, and Claude models built and judged programs. A misreading of the semantics shared by the reference and the Claude-built programs would be invisible to our scoring; independent reproductions from the document text reduce this risk but do not exclude it.

\paragraph{The Scope.} Synthetic workflow data: each run comes from a published simulation model~\cite{chapela2023zenodo}, and every random value it uses is recorded. A real business will not know these values in advance. Questions that come from templates may be easier to interpret than questions written by people. Finally, \Trusted means only that specific tests passed.

\section{Conclusion}
\label{sec:conclusion}

A generated simulator that reproduces recorded history has not shown that it handles the change a what-if question asks about. On programs written by Claude models Haiku 5.5 and Sonnet 5.5, the history check let \rHistWrongClaude{} wrong answers through and Vouch let \rGateWrongClaude, nearly all of the difference from one platform and most of it from changed routing probabilities. Vouch found these errors through failed tests rather than abstention, needed no extra model calls, and at equal coverage accepted fewer wrong answers than LLM judges built on Claude Haiku.

The benefit is conditional. Where the history check already finds the errors, or programs make none or are mostly broken, Vouch adds little, with some correct answers as cost. After a policy change it answered nothing, while agreement with a reading answer kept nearly half of the answers. Open questions are how to recover useful answers once recorded history no longer applies, and whether the gain holds for faults our tests do not target.

\paragraph{Artifact.} The \href{https://anonymous.4open.science/r/vouch_artifact-6893}{\textcolor{ACMDarkBlue}{anonymized artifact}} holds the code, questions, program list, judge prompts, decision rows and analysis scripts; the workflow models are under CC BY 4.0~\cite{chapela2023zenodo}.

\section*{GenAI Usage Disclosure}

Generative AI was used in this work in two ways.

\paragraph{As objects of study.} Claude Sonnet~5.5, Claude Haiku~5.5, gpt-oss-120b, Gemma~4 31B and Qwen3.5~9B built the simulator programs we evaluate. Claude Haiku~5.5 interpreted every question, produced the reading answers, and performed \otf's record edits. These five models also served as the LLM-based judges described in \S\ref{sec:judges}, and TypeSafe's Jev, a hosted decision model that writes no text, served as one further judge. All model calls, tokens, costs and failures are logged in the artifact.

\paragraph{Implementation of the research software and the benchmark.} An AI coding assistant was directed in writing some of the code of the assurance library.

\bibliographystyle{ACM-Reference-Format}
\bibliography{references}

\end{document}